\documentclass[runningheads]{llncs}
\usepackage[T1]{fontenc}
\usepackage{graphicx}
\usepackage{booktabs}
\usepackage[misc]{ifsym}

\usepackage{mwe}
\usepackage{amsmath}
\usepackage{amsfonts}
\usepackage{booktabs}
\usepackage{subcaption}
\usepackage{hyperref}
\usepackage{algorithm}
\usepackage{algorithmic}
\usepackage{placeins}
\usepackage{array}
\usepackage{makecell}
\usepackage{xcolor}
\usepackage{orcidlink}
\definecolor{NoveltyPurple}{HTML}{7B2CBF}
\definecolor{CrossDataBlue}{HTML}{1F77B4}
\definecolor{TemporalOrange}{HTML}{E67E22}
\definecolor{ReproGreen}{HTML}{2E8B57}
\definecolor{ConclusionTeal}{HTML}{008080}
\definecolor{StatsRed}{HTML}{C0392B}
\definecolor{MinorGray}{HTML}{7F8C8D}
\definecolor{HighlightYellow}{HTML}{000000}
\begin{document}

\title{Beyond Frame-Level Accuracy: Analyzing Video-Level Stability in Transformer-Based Deepfake Forensics}

\titlerunning{Analyzing Video-Level Stability in Transformer-Based Deepfake Forensics}

\author{Syed Azmul Hasan\inst{1} \and Mowmita Pervin Hera\inst{1} \and Afroza Lona Mim\inst{1} \and M M Jihad Hossain\inst{1} \and Hazrat Ali\inst{2}\orcidlink{0000-0003-3058-5794} \and Mohammad Farhad Bulbul\inst{1}\orcidlink{0000-0003-1691-8695}
}

\authorrunning{S. A. Hasan et al.}

\institute{Department of Mathematics, Jashore University of Science and Technology, Jashore 7408, Bangladesh\\
\and
School of Computing, Data and Mathematical Sciences, University of Stirling, Stirling, United Kingdom}

\maketitle              

\noindent\textbf{Correspondence:} Mohammad Farhad Bulbul (farhad@just.edu.bd)

\begin{abstract}
Deepfake videos generated by advanced generative models pose serious threats to digital security, misinformation control, and media forensics. Although numerous detection methods have been proposed, their reported performances are often based on heterogeneous preprocessing pipelines, inconsistent data splits, and model-specific training protocols, hindering fair architectural comparison. This work presents a controlled and reproducible comparative study of convolutional neural networks (ResNet-18 and EfficientNet-B0) and a transformer-based architecture (ViT-B/16) within a unified, face-centric deepfake detection framework. All models are trained using identical preprocessing, augmentation, optimization settings, and data splits on Celeb-DF v2 and FaceForensics++, \textcolor{HighlightYellow}{and results are validated across three independent random seeds with bootstrapped confidence intervals}. Experimental results show that while CNN-based models perform competitively at the frame level, their performance degrades after video-level aggregation. In contrast, ViT-B/16 demonstrates more stable video-level performance, \textcolor{HighlightYellow}{achieving 92.8\% [95\% CI: 91.7--94.0] video-level accuracy and an AUC of 0.950 [95\% CI: 0.934--0.964] on Celeb-DF v2 (3-seed bootstrapped)}. \textcolor{HighlightYellow}{Cross-dataset evaluation across the same three random seeds reveals a large, asymmetric generalization gap (AUC 0.79 FF++$\rightarrow$Celeb-DF vs. 0.64 Celeb-DF$\rightarrow$FF++), indicating that in-domain video-level stability does not transfer to unseen manipulation domains.} \textcolor{HighlightYellow}{These findings indicate that global self-attention improves robustness and video-level stability for deepfake detection \emph{within} a given data domain, while highlighting the need for further work on cross-domain generalization.}

\end{abstract}

\section{Introduction}

Recent advances in generative modeling—particularly Generative Adversarial Networks (GANs), autoencoder-based pipelines, and neural face-editing frameworks—have enabled the synthesis of highly realistic facial images and videos, commonly referred to as \textit{deepfakes}. While such technologies have legitimate applications in entertainment and media production, their misuse poses serious risks to digital security, misinformation control, and personal privacy \cite{rossler2019faceforensics}, \cite{dfdc2020}. Continuous improvements in visual realism and temporal coherence have made modern deepfakes increasingly difficult to distinguish from authentic media.

Deepfake detection has therefore emerged as an important research area at the intersection of computer vision and multimedia forensics. Early detection approaches relied on handcrafted artifacts such as eye-blinking inconsistencies and head pose irregularities \cite{li2018ictu}, \cite{yang2019headpose}. However, as generative models improved, these low-level cues became less reliable. Consequently, research shifted toward deep learning-based detectors trained on large-scale manipulated datasets. Benchmarks such as FaceForensics++ and Celeb-DF v2 have enabled standardized evaluation and accelerated progress in this field \cite{rossler2019faceforensics}, \cite{li2020celebdf}.

Despite substantial advances, several challenges remain. Many detectors exhibit limited cross-dataset generalization and performance degradation under compression or unseen manipulation methods \cite{ojha2023universal}. Furthermore, most existing approaches rely on Convolutional Neural Networks (CNNs), including EfficientNet and ResNet variants \cite{tan2019efficientnet}, \cite{he2016resnet}. While CNNs effectively capture local texture artifacts, their limited receptive fields restrict their ability to model long-range spatial dependencies. Prior work suggests that incorporating global context can improve robustness to subtle and spatially distributed manipulations \cite{heo2021deepfake}, \cite{wang2022m2tr}.

Vision Transformers (ViTs) have demonstrated competitive or superior performance compared to CNNs across various visual recognition tasks \cite{dosovitskiy2021vit}. In deepfake detection, global self-attention may enhance robustness by integrating evidence across multiple facial regions rather than focusing solely on localized texture inconsistencies. However, many transformer-based detectors are evaluated under different preprocessing pipelines and training protocols, making a fair architectural comparison difficult.

Motivated by these limitations, this paper presents a controlled backbone-level comparison between CNN architectures (EfficientNet-B0 and ResNet-18) and a transformer-based model (ViT-B/16) for deepfake detection. All models are trained within a unified face-centric pipeline using identical preprocessing, augmentation, data splits, and optimization settings on Celeb-DF v2 \cite{li2020celebdf} and FaceForensics++ \cite{rossler2019faceforensics}. By eliminating experimental inconsistencies, we isolate architectural differences and provide a fair comparative evaluation.

Beyond frame-level analysis, we investigate the impact of video-level aggregation by averaging frame-wise prediction scores to obtain stable video-level decisions. We further analyze model interpretability using Grad-CAM for CNNs and attention visualization for transformers to examine spatial activation patterns.

The main contributions of this work are:

\begin{itemize}
    \item We present a controlled and reproducible backbone-level comparison of CNN and transformer architectures under identical experimental conditions.
    \item We provide a systematic analysis of frame-level versus video-level performance and aggregation stability.
    \item We provide a comparative interpretability study highlighting differences in spatial activation patterns.
    \item We provide empirical insights into the robustness and computational trade-offs for practical deployment.
    \item \textcolor{HighlightYellow}{We provide a 3-seed, statistically validated cross-dataset generalization study, quantifying the extent to which video-level stability transfers across domains.}
    \item \textcolor{HighlightYellow}{We empirically test an explicit temporal-modeling alternative (a Gated Recurrent Unit, GRU-based aggregation) against mean pooling, reporting a preliminary negative result that motivates directions for future work.}
\end{itemize}

The remainder of this paper is organized as follows. Section 2 reviews prior work on deepfake generation and detection, covering CNN-based, frequency-domain, temporal, and transformer-based approaches. Section 3 presents the proposed unified face-centric detection framework, detailing preprocessing, backbone architectures, and training protocol. Section 4 reports quantitative and qualitative experimental results, including frame-level and video-level evaluations, \textcolor{HighlightYellow}{statistical robustness}, and \textcolor{HighlightYellow}{cross-dataset generalization}. Section 5 discusses the impact of global self-attention, aggregation behavior, interpretability, computational trade-offs, and \textcolor{HighlightYellow}{limitations}. Finally, Section 6 concludes the paper and outlines potential directions for future research.

\section{Related Work}
The rapid advancement of generative models, such as Generative Adversarial Networks (GANs) and neural rendering techniques, has enabled the creation of highly realistic manipulated videos, commonly referred to as deepfakes. While these technologies offer promising applications in entertainment and media production, they introduce severe risks regarding misinformation and digital security. Consequently, extensive research has shifted toward developing robust, automated detection methods leveraging computer vision and deep learning.

\subsection{Artifact-Based Spatial Detection}

Early detection methods primarily focused on identifying visual artifacts introduced during the synthesis process, such as inconsistencies in blending, boundary generation, and texture synthesis. The field saw substantial progress with the introduction of large-scale benchmarks, particularly FaceForensics++ by R\"ossler et al.~\cite{rossler2019faceforensics}. This study demonstrated that deep convolutional neural networks (CNNs), specifically XceptionNet, could achieve high accuracy by treating deepfake detection as a per-frame classification task.

Building on this, researchers explored more sophisticated architectures to enhance feature representation. Lacerda et al.~\cite{lacerda2022efficientnet} utilized EfficientNet-based models to improve performance through compound scaling and parameter efficiency. Meanwhile, Wang et al.~\cite{wang2020cnn} conducted a comprehensive study on CNN-based detectors, highlighting the significant challenges of generalization across different generative models. Alternative architectures like Capsule-Forensics~\cite{nguyen2019capsule} were also proposed to preserve hierarchical relationships between facial features. To target specific synthesis flaws, Li et al.~\cite{li2020facexray} introduced Face X-ray, which explicitly models blending boundaries to identify manipulation without prior knowledge of the synthesis method.

\subsection{Spatiotemporal and Physiological Approaches}

While spatial methods are effective at capturing frame-level artifacts, they often overlook temporal inconsistencies. To address this, researchers integrated temporal dynamics and physiological cues. Qi et al.~\cite{deepRhythm} introduced DeepRhythm, which utilizes remote photoplethysmography (rPPG) to detect subtle, heartbeat-induced variations in facial regions—a biological signal that generative models struggle to replicate. Similarly, Prashnani et al.~\cite{prashnani2022phaseforensics} proposed PhaseForensics, leveraging phase-based motion representations and 3D CNNs to capture temporal irregularities in motion that are imperceptible in the spatial domain.

Furthermore, lightweight alternatives have been explored for computational efficiency. Chen et al.~\cite{chen2021defakehop} developed DefakeHop, a framework based on successive subspace learning. Unlike standard deep networks, DefakeHop extracts hierarchical features through a data-driven approach without backpropagation, maintaining competitive performance on benchmarks like Celeb-DF~\cite{li2020celebdf} while requiring significantly fewer parameters.

\subsection{Attention and Multi-Domain Feature Learning}

To improve the localization of subtle artifacts, recent studies have incorporated attention mechanisms. Zhao et al.~\cite{dang2020detection} proposed a multi-attentional framework using multiple spatial attention heads to capture artifacts across various facial regions simultaneously. Similarly, Omar et al.~\cite{zhang2023selfattention} utilized self-attention mechanisms to dynamically emphasize discriminative spatial patterns, improving accuracy in high-quality, low-artifact scenarios.

Beyond attention, multi-domain learning has gained traction. Xu et al.~\cite{xu2023mcxapi} introduced MCX-API, which integrates multi-channel color-space representations with pairwise interaction learning. By capturing cross-channel dependencies, this approach enhances the model's ability to generalize across diverse datasets, bridging the performance gap between FaceForensics++ and Celeb-DF.

\subsection{Benchmark Datasets and Evaluation}

The evolution of detection algorithms has been driven by benchmark datasets. FaceForensics++ \cite{rossler2019faceforensics} remains the standard for controlled manipulation studies, while Celeb-DF v2 \cite{li2020celebdf} provides a more challenging and high-quality benchmark with fewer visual artifacts. Due to the high realism of these datasets, performance is standardly evaluated using the Area Under the ROC Curve (AUC) to ensure a threshold-independent assessment. 

\subsection{Summary and Research Gap}

Despite the progress in CNN-based and spatiotemporal methods, two primary challenges remain. Most CNNs overfit to specific dataset artifacts and fail when tested on unseen manipulation techniques. In addition, traditional architectures often focus on local textures and fail to capture global structural dependencies across the entire face.

To address these limitations, recent research has shifted toward Transformer-based architectures, which utilize self-attention to capture long-range dependencies and global contextual relationships. Motivated by these advantages, this work proposes a unified framework built on the ViT-B/16 architecture. By employing a consistent preprocessing pipeline and leveraging the global receptive field of Transformers, the proposed method aims to achieve superior robustness and cross-dataset generalization on both the Celeb-DF and FaceForensics++ benchmarks.

\section{Methodology}
The proposed pipeline consists of face-centric preprocessing, frame-level classification, and video-level aggregation under identical experimental settings. In addition, this section describes \textcolor{HighlightYellow}{the cross-dataset evaluation protocol (\S 3.5)}, \textcolor{HighlightYellow}{the temporal aggregation ablation (\S 3.6)}, and \textcolor{HighlightYellow}{the statistical evaluation procedure (\S 3.7)} used to validate all reported results.

\subsection{System Overview}
\begin{figure}[t]
    \centering
\includegraphics[width=0.9\linewidth]{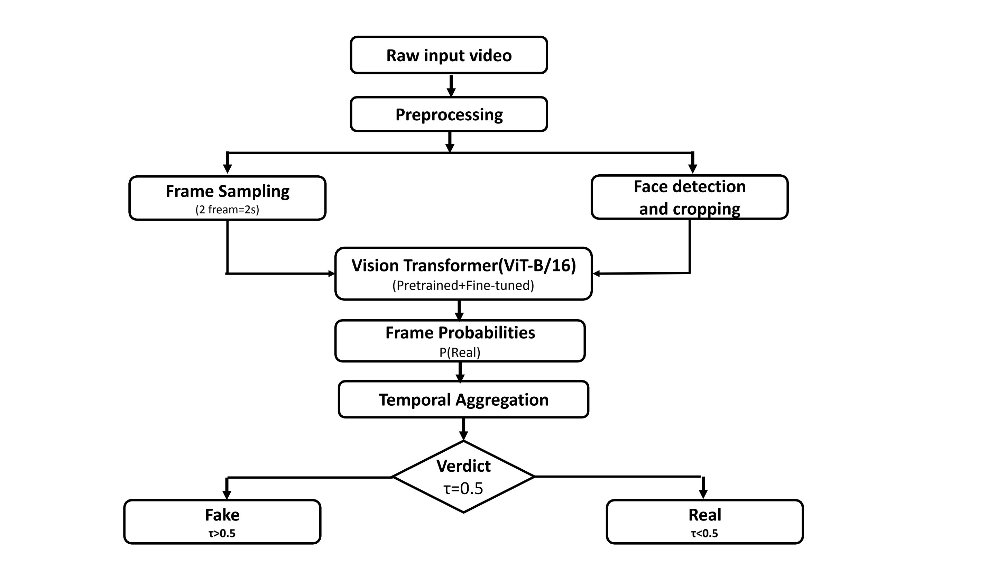}
    \caption{Overall workflow of the proposed deepfake detection pipeline.}
    \label{fig:flowchart}
\end{figure}

Given an input video, frames are uniformly sampled at fixed temporal intervals to ensure consistent coverage while maintaining computational efficiency. Each sampled frame is processed using MTCNN for face detection, and the largest detected face is cropped and resized to $224 \times 224$.
The cropped region of the face is passed to a backbone network (ResNet-18, EfficientNet-B0, or ViT-B/16), which outputs a frame-level manipulation probability $p_i$. For a video containing $K$ sampled frames, predictions are aggregated using mean pooling:
\begin{equation}
P_{\text{video}} = \frac{1}{K} \sum_{k=1}^{K} p_k,
\end{equation}
where $P_{\text{video}}$ denotes the final video-level prediction. A threshold $\tau = 0.5$ is used for binary classification. Figure \ref{fig:flowchart} shows the overall workflow of the pipeline.

\subsection{Datasets and Preprocessing}
Experiments were conducted on Celeb-DF v2 \cite{li2020celebdf} and FaceForensics++ \cite{rossler2019faceforensics}, two widely adopted deepfake benchmarks. Both datasets are split at the video level into 80\% training, 10\% validation, and 10\% testing sets to prevent data leakage. \textcolor{HighlightYellow}{Exact split sizes and per-split real/fake frame counts are reported in Table \ref{tab:splits}.}

Frames are uniformly sampled from each video. Detected faces are resized to $224 \times 224$ and normalized using ImageNet statistics. Data augmentation includes random resized cropping, horizontal flipping, color jitter, and optional Gaussian blur or noise to improve robustness against compression artifacts. \textcolor{HighlightYellow}{We note that both datasets are class-imbalanced and skewed toward the fake class (Celeb-DF v2: 84--87\% fake frames; FaceForensics++: 65--69\% fake frames), and accuracy figures reported in Section 4 should therefore be interpreted alongside F1 and AUC rather than in isolation.}

\begin{table}[t]
\centering
\caption{\textcolor{HighlightYellow}{Dataset split statistics, including class balance.}}
\label{tab:splits}
\begin{tabular}{lllllll}
\hline
Dataset & Split & Videos & Real frames & Fake frames & Total frames & Fake ratio \\
\hline
FF++ & Train & 5{,}600 & 71{,}545 & 136{,}784 & 208{,}329 & 65.7\% \\
FF++ & Val & 700 & 9{,}078 & 17{,}065 & 26{,}143 & 65.3\% \\
FF++ & Test & 700 & 8{,}157 & 18{,}164 & 26{,}321 & 69.0\% \\
Celeb-DF v2 & Train & 5{,}223 & 6{,}901 & 45{,}320 & 52{,}221 & 86.8\% \\
Celeb-DF v2 & Val & 653 & 1{,}040 & 5{,}490 & 6{,}530 & 84.1\% \\
Celeb-DF v2 & Test & 653 & 950 & 5{,}580 & 6{,}530 & 85.5\% \\
\hline
\end{tabular}
\end{table}

\subsection{Model Architectures}
Three backbone architectures were evaluated under identical preprocessing and optimization settings:

\textbf{CNN-based models.} ResNet-18 and EfficientNet-B0 were initialized with ImageNet-pretrained weights and fine-tuned for binary classification.

\begin{figure*}[t]
\centering
\includegraphics[width=\textwidth]{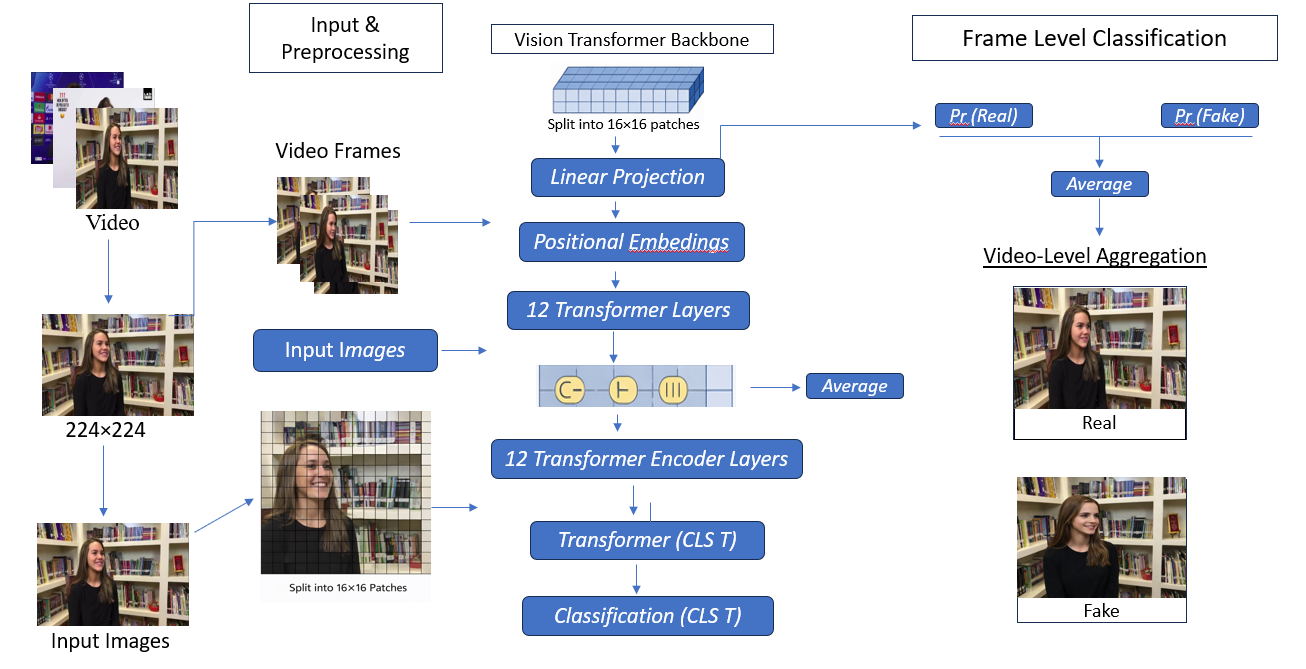}
\caption{Architecture of ViT-B/16 used for deepfake detection.}
\label{fig:vit-arch}
\end{figure*}

\textbf{Transformer-based model.} ViT-B/16 partitions each input image into $16 \times 16$ patches and processes them using 12 transformer encoder layers with multi-head self-attention. The final classification token is used to predict manipulation probability. Figure \ref{fig:vit-arch} shows the architecture summary of the ViT model used in this work.
\begin{figure}[tbh!]
\centering
\includegraphics[width=0.8\textwidth]{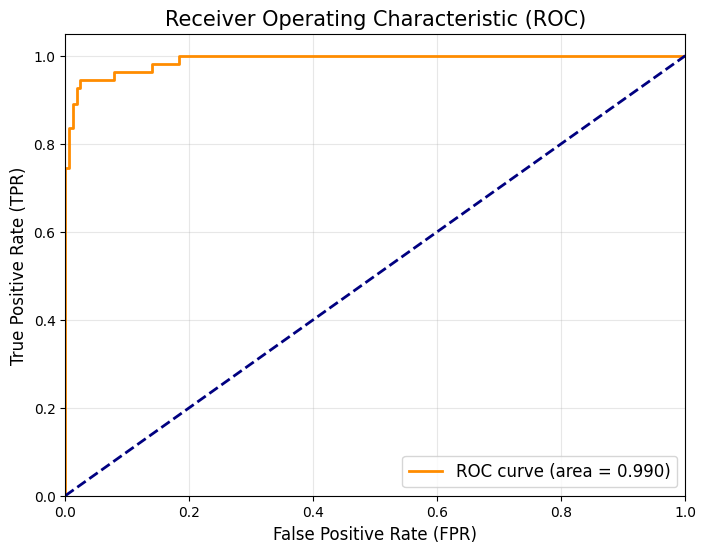}
\caption{ROC curve of ViT-B/16 on FaceForensics++.}
\label{fig:roc-curve1}
\end{figure}
\begin{figure}[tbh!]
\centering
\includegraphics[width=0.8\textwidth]{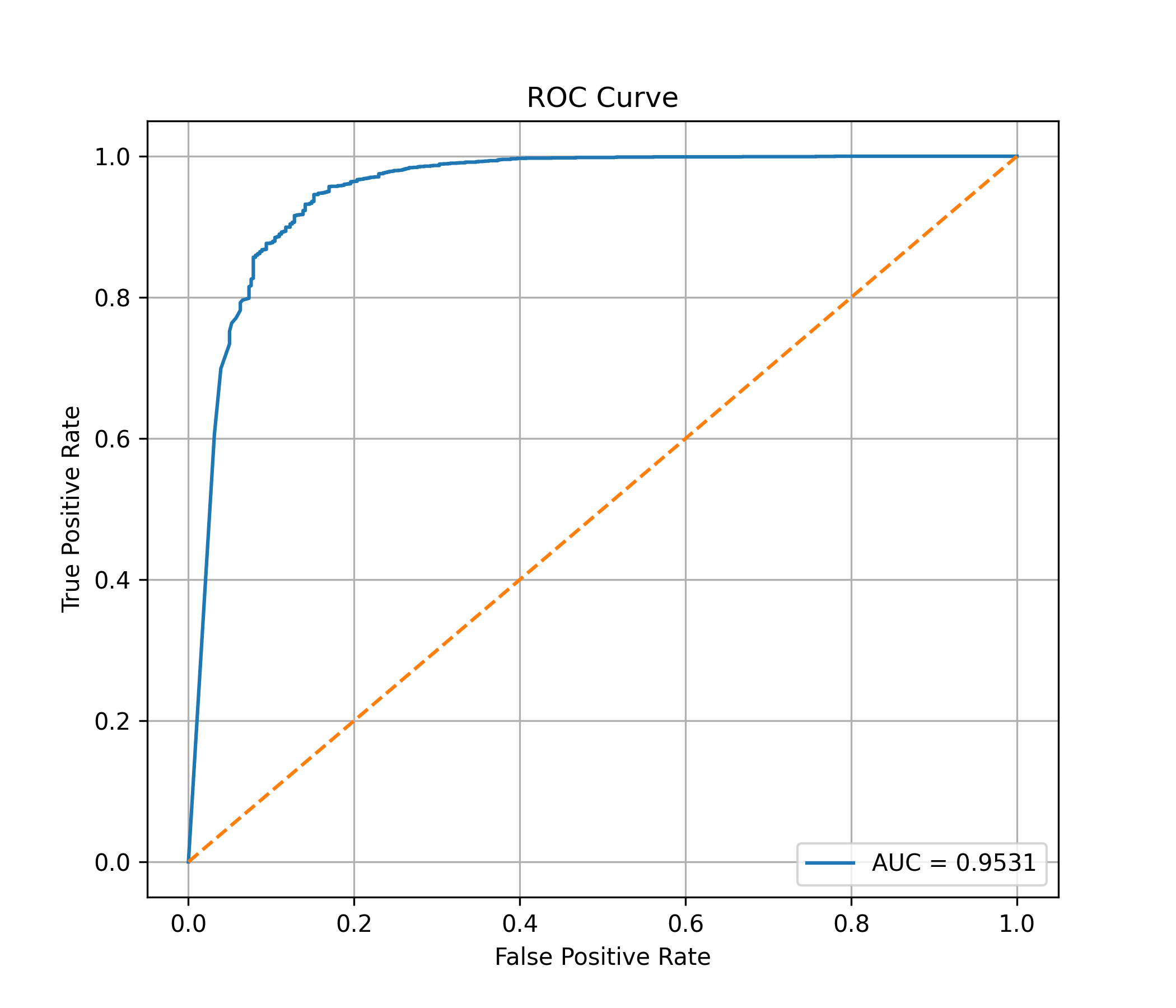}
\caption{ROC curve of ViT-B/16 on Celeb-DF v2.}
\label{fig:roc-curve_df}
\end{figure}

\subsection{Training Protocol}
\textcolor{HighlightYellow}{All models are trained using binary cross-entropy loss and the AdamW optimizer with layer-wise learning-rate decay of 0.7 and a base learning rate of $1.4\times10^{-5}$, weight decay of 0.06, and an effective batch size of 32 (micro-batch size 16 with gradient accumulation of two). Models are trained for 10 epochs on FaceForensics++ and 6 epochs on Celeb-DF v2, with the checkpoint selected by best exponential-moving-average (EMA) validation ROC-AUC. All experiments were run on a single Tesla T4 GPU (14.56 GB VRAM) using PyTorch 2.10.0+cu128 and timm 1.0.26.}

To ensure fair comparison, identical data splits, augmentation strategies, and evaluation metrics were used across all architectures. \textcolor{HighlightYellow}{To assess reproducibility and seed sensitivity, every configuration (both datasets, both cross-dataset directions, and the mean-pooling backbone) was trained and evaluated independently under three random seeds (42, 123, 2024);} \textcolor{HighlightYellow}{results are aggregated and reported with confidence intervals in Section 4, as detailed in \S 3.7.} \textcolor{HighlightYellow}{Full hyperparameter and hardware settings are summarized in Table \ref{tab:hyperparams} for reproducibility.}

\begin{table}[t]
\centering
\caption{\textcolor{HighlightYellow}{Reproducibility and hyperparameter settings.}}
\label{tab:hyperparams}
\small
\renewcommand{\arraystretch}{1.12}
\setlength{\tabcolsep}{5pt}

\begin{tabular}{p{3.0cm}p{6.5cm}}
\toprule
\textbf{Item} & \textbf{Value} \\
\midrule

\textcolor{HighlightYellow}{Hardware} &
\textcolor{HighlightYellow}{1$\times$ Tesla T4 GPU, 14.56 GB VRAM} \\

\textcolor{HighlightYellow}{Software} &
\textcolor{HighlightYellow}{PyTorch 2.10.0+cu128, timm 1.0.26} \\

\textcolor{HighlightYellow}{Random seeds} &
\textcolor{HighlightYellow}{42, 123, 2024 (train + in-domain evaluation + both cross-dataset directions)} \\

\textcolor{HighlightYellow}{Data split} &
\textcolor{HighlightYellow}{80\%/10\%/10\% (train/validation/test), split at video level} \\

Image resolution &
$224 \times 224$ \\

\textcolor{HighlightYellow}{Batch size} &
\textcolor{HighlightYellow}{16 (micro-batch) $\times$ 2 (gradient accumulation) = 32 effective} \\

\textcolor{HighlightYellow}{Epochs (mean-pool)} &
\textcolor{HighlightYellow}{10 (FF++), 6 (Celeb-DF v2)} \\

\textcolor{HighlightYellow}{Epochs (temporal GRU)} &
\textcolor{HighlightYellow}{8 (FF++ only, single seed, exploratory)} \\

\textcolor{HighlightYellow}{Optimizer} &
\textcolor{HighlightYellow}{AdamW, layer-wise learning-rate decay = 0.7} \\

\textcolor{HighlightYellow}{Base learning rate} &
\textcolor{HighlightYellow}{$1.4\times10^{-5}$ (backbone); $1\times10^{-4}$ (temporal GRU head)} \\

\textcolor{HighlightYellow}{Weight decay} &
\textcolor{HighlightYellow}{0.06} \\

\textcolor{HighlightYellow}{Checkpoint selection} &
\textcolor{HighlightYellow}{Best EMA validation AUC} \\

\textcolor{HighlightYellow}{Temporal head} &
\textcolor{HighlightYellow}{GRU, hidden size 256, sequence length 16 frames} \\

\textcolor{HighlightYellow}{Statistical testing} &
\textcolor{HighlightYellow}{Bootstrap resampling (2000 iterations), 95\% confidence interval, pooled across three seeds} \\

\bottomrule
\end{tabular}
\end{table}

\textcolor{HighlightYellow}{
\subsection{Cross-Dataset Evaluation Protocol}
\label{subsec:cross-dataset}
To evaluate generalization beyond the training domain, we perform cross-dataset testing by evaluating each trained model zero-shot on the held-out test split of the other dataset, without any fine-tuning or adaptation. Specifically, models trained on the full FaceForensics++ training split are evaluated on the Celeb-DF v2 test split (FF++$\rightarrow$Celeb-DF), and vice versa (Celeb-DF$\rightarrow$FF++). The same face-centric preprocessing pipeline (MTCNN detection, $224\times224$ cropping, ImageNet normalization) is applied in both directions, with no dataset-specific tuning. For evaluation compute efficiency, a frame cap of 15 frames per video is used during cross-dataset inference. This protocol is repeated independently for all three random seeds in both directions, yielding six cross-dataset evaluation runs in total. Results are reported in \S 4.1.
}

\textcolor{HighlightYellow}{
\subsection{Temporal Aggregation Ablation}
\label{subsec:temporal-ablation}
Video-level aggregation in the main pipeline (Eq. 1) uses simple mean pooling over frame-level probabilities, which does not explicitly model temporal dependencies between frames. To test whether explicit temporal modeling improves upon this simple aggregation strategy, we implement a Gated Recurrent Unit (GRU) \cite{cho2014gru}-based temporal head as an alternative aggregator. The GRU operates on frozen frame-level embeddings extracted from the ViT-B/16 backbone, using a sequence length of 16 frames per video, a hidden size of 256, and is trained with the AdamW optimizer at a learning rate of $1\times10^{-4}$ for 8 epochs. This ablation was run on FaceForensics++ using a single random seed (42) due to time constraints, and is reported as a preliminary, exploratory result rather than a fully validated finding; it was not repeated across all three seeds or on Celeb-DF v2. Results are reported in \S 4.1 and discussed in \S 5.2.
}

\textcolor{HighlightYellow}{
\subsection{Statistical Evaluation}
\label{subsec:stats}
All in-domain and cross-dataset results are additionally evaluated using bootstrap resampling. For each experimental condition (in-domain FF++, in-domain Celeb-DF v2, and both cross-dataset directions), video-level predictions from all three completed random seeds are pooled, and 2000 bootstrap iterations are performed by resampling videos with replacement to compute 95\% confidence intervals for Accuracy, F1 score, and ROC-AUC. This procedure allows us to report point estimates alongside confidence intervals (Table~\ref{tab:bootstrap_indomain} for in-domain results and Table~\ref{tab:cross_bootstrap} for cross-dataset results), demonstrating statistical significance analysis and seed-stability evidence.
}

\subsection{Evaluation Metrics}
Performance is evaluated at both frame and video levels using Accuracy, F1 score, and ROC-AUC. Video-level predictions are obtained via mean aggregation of frame-level probabilities, \textcolor{HighlightYellow}{except where explicitly noted otherwise (\S 3.6)}. \textcolor{HighlightYellow}{See \S 3.7 for the statistical evaluation methodology underlying the confidence intervals reported alongside these metrics in Section 4.}


\section{Experimental Results}

\subsection{Quantitative Evaluation}

First, we evaluated the comparative performance of CNN-based baselines and the proposed ViT-B/16.
Table~\ref{tab:vit_details} summarizes the configuration of the ViT-B/16 model used in all experiments.

\begin{table}[t]
\centering
\caption{Configuration of ViT-B/16.}
\label{tab:vit_details}
\small
\begin{tabular}{l l}
\toprule
Parameter & Specification \\
\midrule
Patch Size & $16 \times 16$ \\
Embedding Dimension & 768 \\
Layers & 12 \\
Attention Heads & 12 \\
Dropout & 0.1 \\
Pretraining & ImageNet-1k \\
Optimizer & AdamW (layer-wise LR decay = 0.7) \\
Base Learning Rate & 1.4e-5 \\
Weight Decay & 0.06 \\
Batch Size & 32 (effective; 16 $\times$ grad. accum. 2) \\
Resolution & $224 \times 224$ \\
\bottomrule
\end{tabular}
\end{table}

\textcolor{HighlightYellow}{To demonstrate statistical robustness, all ViT-B/16 results below were obtained across three independent random seeds (42, 123, 2024) and are reported as bootstrapped estimates (2000 resampling iterations, 95\% CI, pooled across seeds; see \S 3.7). CNN baselines (ResNet-18, EfficientNet-B0) were trained once under the same data splits and preprocessing and are reported as single-run results; the primary purpose of the multi-seed protocol is to validate the stability of the proposed transformer-based pipeline, not to re-benchmark the baselines.} Video-level performance results are presented in Table~\ref{tab:model_comparison_summary}. \textcolor{HighlightYellow}{Frame-level computational performance is reported in Table~\ref{tab:vit_latency_summary}, including peak GPU memory usage measured separately during a single forward (inference) pass and during a full forward-backward (training) step, both including face-detection preprocessing overhead.}

\begin{table*}[t]
\centering
\caption{\textcolor{HighlightYellow}{Video-level performance comparison on Celeb-DF v2 and FaceForensics++ (FF++). CNN baselines are single-run; ViT-B/16 rows report the 3-seed bootstrapped mean with 95\% CI (2000 iterations).}}
\label{tab:model_comparison_summary}
\small
\renewcommand{\arraystretch}{1.15}
\setlength{\tabcolsep}{4pt}

\begin{tabular}{l l r c c c c}
\toprule
\textbf{Model} &
\textbf{Dataset} &
\textbf{Params} &
\textbf{Acc (\%)} &
\textbf{F1} &
\textbf{AUC} &
\textbf{FPS} \\
\midrule

EfficientNet-B0 &
Celeb-DF &
5.30M &
60.71 &
0.670 &
0.763 &
18.3 \\

ResNet-18 &
Celeb-DF &
11.18M &
88.04 &
0.907 &
0.949 &
270.3 \\

ViT-B/16 &
Celeb-DF &
85.8M &
\makecell[c]{\textbf{\textcolor{HighlightYellow}{92.83}}\\{\scriptsize \textcolor{HighlightYellow}{[91.68, 93.98]}}} &
\makecell[c]{\textbf{\textcolor{HighlightYellow}{0.958}}\\{\scriptsize \textcolor{HighlightYellow}{[0.951, 0.965]}}} &
\makecell[c]{\textbf{\textcolor{HighlightYellow}{0.950}}\\{\scriptsize \textcolor{HighlightYellow}{[0.934, 0.964]}}} &
18.3 \\

\midrule

EfficientNet-B0 &
FF++ &
5.30M &
80.22 &
0.895 &
0.913 &
5.09 \\

ResNet-18 &
FF++ &
11.18M &
82.00 &
0.837 &
0.955 &
15.8 \\

ViT-B/16 &
FF++ &
85.8M &
\makecell[c]{\textbf{\textcolor{HighlightYellow}{77.41}}\\{\scriptsize \textcolor{HighlightYellow}{[75.57, 79.10]}}} &
\makecell[c]{\textbf{\textcolor{HighlightYellow}{0.835}}\\{\scriptsize \textcolor{HighlightYellow}{[0.821, 0.849]}}} &
\makecell[c]{\textbf{\textcolor{HighlightYellow}{0.840}}\\{\scriptsize \textcolor{HighlightYellow}{[0.821, 0.859]}}} &
18.3 \\

\bottomrule
\end{tabular}
\end{table*}

\begin{table}[t]
\centering
\caption{\textcolor{HighlightYellow}{Frame-level latency and GPU memory usage of ViT-B/16 (Tesla T4, 14.56 GB).}}
\label{tab:vit_latency_summary}
\small
\begin{tabular}{l l l l l}
\toprule
Dataset & Latency (ms) & FPS & \textcolor{HighlightYellow}{Peak GPU Mem., Inference (GB)} & \textcolor{HighlightYellow}{Peak GPU Mem., Training (GB)} \\
\midrule
Celeb-DF & 13.57 & 73.70 & \textcolor{HighlightYellow}{0.668} & \textcolor{HighlightYellow}{2.835--2.839} \\
FF++ & 11.80 & 84.75 & \textcolor{HighlightYellow}{0.668} & \textcolor{HighlightYellow}{2.837--2.839} \\
\bottomrule
\end{tabular}
\end{table}

\textcolor{HighlightYellow}{
\subsubsection{Statistical Robustness}
\label{sssec:stat-robustness}
To provide statistical validation of the reported results (\S 3.7), Table~\ref{tab:seed_breakdown} reports the raw per-seed accuracy, F1, and AUC values obtained independently under each of the three random seeds; these three per-seed runs are the underlying data that were pooled and bootstrapped to produce the ViT-B/16 summary rows in Table~\ref{tab:model_comparison_summary}. Table~\ref{tab:bootstrap_indomain} reports the resulting bootstrapped point estimates and 95\% confidence intervals in full. All three seeds completed the full in-domain training and evaluation pipeline. The seed-to-seed spread is small (video-level AUC: 0.8389--0.8416 on FF++, 0.9477--0.9539 on Celeb-DF v2), and the resulting bootstrapped confidence intervals are correspondingly tight, which indicate that the proposed pipeline's video-level performance is reproducible and not an artifact of a single favorable training run.
}

\begin{table}[t]
\centering
\caption{\textcolor{HighlightYellow}{In-domain results per seed for ViT-B/16.}}
\label{tab:seed_breakdown}
\small
\begin{tabular}{l r r r r r}
\toprule
Dataset & Seed & Frame AUC & Video AUC & Video Acc & Video F1 \\
\midrule
FF++ & 42 & 0.8140 & 0.8416 & 0.7800 & 0.8392 \\
FF++ & 123 & 0.8102 & 0.8409 & 0.7757 & 0.8370 \\
FF++ & 2024 & 0.8074 & 0.8389 & 0.7671 & 0.8304 \\
Celeb-DF v2 & 42 & 0.9276 & 0.9539 & 0.9326 & 0.9606 \\
Celeb-DF v2 & 123 & 0.9203 & 0.9477 & 0.9280 & 0.9577 \\
Celeb-DF v2 & 2024 & 0.9229 & 0.9486 & 0.9250 & 0.9557 \\
\bottomrule
\end{tabular}
\end{table}

\begin{table}[t]
\centering
\caption{\textcolor{HighlightYellow}{In-domain results, bootstrapped (2000 iterations, 95\% CI, pooled across all 3 seeds).}}
\label{tab:bootstrap_indomain}
\small
\renewcommand{\arraystretch}{1.15}
\setlength{\tabcolsep}{4pt}

\begin{tabular}{l r c c c}
\toprule
\textbf{Dataset} &
\textbf{N videos} &
\makecell[c]{\textbf{Acc}\\\textbf{(mean [95\% CI])}} &
\makecell[c]{\textbf{F1}\\\textbf{(mean [95\% CI])}} &
\makecell[c]{\textbf{AUC}\\\textbf{(mean [95\% CI])}} \\
\midrule

FF++ &
2100 &
\makecell[c]{\textcolor{HighlightYellow}{0.7741}\\{\scriptsize \textcolor{HighlightYellow}{[0.7557, 0.7910]}}} &
\makecell[c]{\textcolor{HighlightYellow}{0.8353}\\{\scriptsize \textcolor{HighlightYellow}{[0.8207, 0.8489]}}} &
\makecell[c]{\textcolor{HighlightYellow}{0.8404}\\{\scriptsize \textcolor{HighlightYellow}{[0.8206, 0.8586]}}} \\

Celeb-DF v2 &
1959 &
\makecell[c]{\textcolor{HighlightYellow}{0.9283}\\{\scriptsize \textcolor{HighlightYellow}{[0.9168, 0.9398]}}} &
\makecell[c]{\textcolor{HighlightYellow}{0.9578}\\{\scriptsize \textcolor{HighlightYellow}{[0.9507, 0.9647]}}} &
\makecell[c]{\textcolor{HighlightYellow}{0.9499}\\{\scriptsize \textcolor{HighlightYellow}{[0.9340, 0.9643]}}} \\

\bottomrule
\end{tabular}
\end{table}

\begin{figure}[tbh!]
\centering
\includegraphics[width=0.7\textwidth]{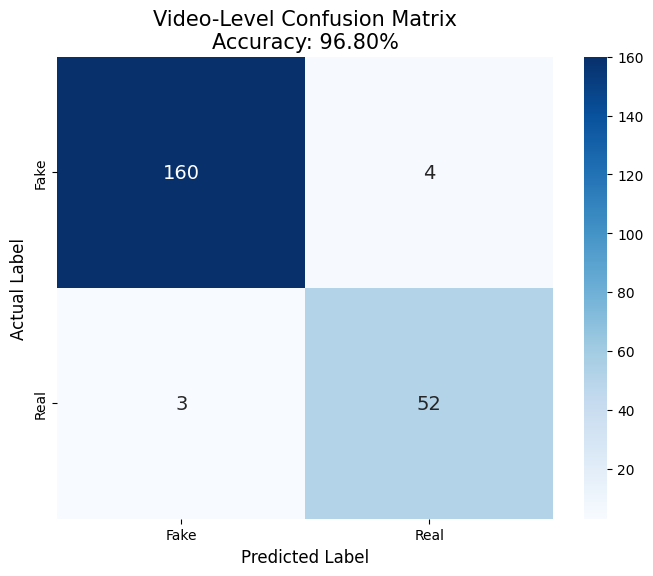}
\caption{Video-level confusion matrix for ViT-B/16 on the FF++ test set \textcolor{HighlightYellow}{(representative single seed, seed 42, $n=700$ videos; see Table~\ref{tab:bootstrap_indomain} for the 3-seed bootstrapped aggregate)}.}
\label{fig:conf-matrix}
\end{figure}

\begin{figure}[tbh!]
\centering
\includegraphics[width=0.7\textwidth]{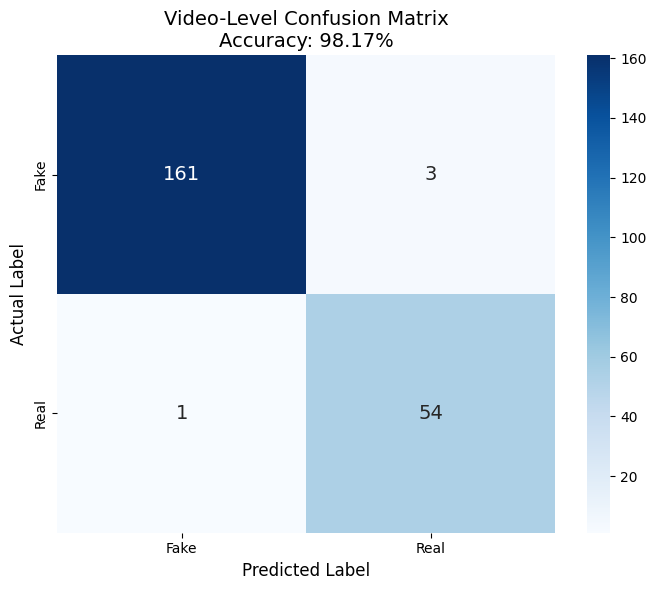}
\caption{Video-level confusion matrix for ViT-B/16 on the Celeb-DF v2 test set \textcolor{HighlightYellow}{(representative single seed, seed 42, $n=653$ videos; see Table~\ref{tab:bootstrap_indomain} for the 3-seed bootstrapped aggregate)}.}
\label{fig:conf-matrix_df}
\end{figure}

\textcolor{HighlightYellow}{
\subsubsection{Cross-Dataset Generalization}
\label{sssec:cross-dataset}
Following the cross-dataset protocol described in \S 3.5, we evaluate each trained model zero-shot on the other dataset's held-out test split, repeated across all three random seeds in both directions. Per-seed results are shown in Table~\ref{tab:cross_perseed}, and the corresponding bootstrapped aggregates are shown in Table~\ref{tab:cross_bootstrap}.
}

\begin{table}[t]
\centering
\caption{\textcolor{HighlightYellow}{Cross-dataset generalization results per seed.}}
\label{tab:cross_perseed}
\small
\begin{tabular}{l r r r r r}
\toprule
Direction & Seed & Frame AUC & Video AUC & Video Acc & Video F1 \\
\midrule
FF++ $\rightarrow$ Celeb-DF & 42 & 0.7352 & 0.7752 & 0.6570 & 0.7586 \\
FF++ $\rightarrow$ Celeb-DF & 123 & 0.7531 & 0.7956 & 0.6708 & 0.7705 \\
FF++ $\rightarrow$ Celeb-DF & 2024 & 0.7571 & 0.7967 & 0.6815 & 0.7783 \\
Celeb-DF $\rightarrow$ FF++ & 42 & 0.6338 & 0.6475 & 0.3929 & 0.3222 \\
Celeb-DF $\rightarrow$ FF++ & 123 & 0.6189 & 0.6301 & 0.4000 & 0.3375 \\
Celeb-DF $\rightarrow$ FF++ & 2024 & 0.6269 & 0.6409 & 0.3700 & 0.2782 \\
\bottomrule
\end{tabular}
\end{table}

\begin{table}[t]
\centering
\caption{\textcolor{HighlightYellow}{Cross-dataset generalization, bootstrapped (2000 iterations, 95\% CI, pooled across all 3 seeds).}}
\label{tab:cross_bootstrap}
\small
\renewcommand{\arraystretch}{1.2}
\setlength{\tabcolsep}{5pt}
\begin{tabular}{l r c c c}
\toprule
\textbf{Direction} &
\textbf{N videos} &
\textbf{Accuracy} &
\textbf{F1} &
\textbf{AUC} \\
\midrule
FF++ $\rightarrow$ Celeb-DF v2 &
1959 &
\makecell[c]{\textcolor{HighlightYellow}{0.6691}\\{\footnotesize \textcolor{HighlightYellow}{[0.6483, 0.6901]}}} &
\makecell[c]{\textcolor{HighlightYellow}{0.7685}\\{\footnotesize \textcolor{HighlightYellow}{[0.7505, 0.7861]}}} &
\makecell[c]{\textcolor{HighlightYellow}{0.7881}\\{\footnotesize \textcolor{HighlightYellow}{[0.7617, 0.8133]}}} \\
Celeb-DF v2 $\rightarrow$ FF++ &
2100 &
\makecell[c]{\textcolor{HighlightYellow}{0.3877}\\{\footnotesize \textcolor{HighlightYellow}{[0.3671, 0.4100]}}} &
\makecell[c]{\textcolor{HighlightYellow}{0.3133}\\{\footnotesize \textcolor{HighlightYellow}{[0.2851, 0.3407]}}} &
\makecell[c]{\textcolor{HighlightYellow}{0.6386}\\{\footnotesize \textcolor{HighlightYellow}{[0.6141, 0.6642]}}} \\
\bottomrule
\end{tabular}
\end{table}

The generalization gap is large and asymmetric, and it is confirmed stable across three independent seeds: the confidence intervals in Table~\ref{tab:cross_bootstrap} do not overlap with the corresponding in-domain intervals in Table~\ref{tab:bootstrap_indomain} at all. Models trained on FaceForensics++ retain moderate discriminative power when evaluated on Celeb-DF v2 (AUC $\approx 0.79$), whereas models trained on Celeb-DF v2 exhibit a marked drop in accuracy when evaluated on FaceForensics++ (38.8\% [36.7, 41.0]), falling below the FF++ test set's majority-class baseline of 69.0\% (Table~\ref{tab:splits}). Notably, this 38.8\% accuracy figure in the Celeb-DF$\rightarrow$FF++ direction is not accompanied by an equivalent collapse in ranking ability: AUC in that same direction remains at 0.639 [0.614, 0.664], meaningfully above the chance level of 0.5. This dissociation between a moderately-preserved AUC and a severely degraded fixed-threshold accuracy is characteristic of a \emph{calibration} failure under domain shift rather than a wholesale loss of discriminative signal: the classification threshold ($\tau = 0.5$) is implicitly calibrated to the score distribution of the training domain, and when the model is applied zero-shot to a dataset with a different manipulation distribution and class balance, its output probabilities shift in a way that the fixed threshold no longer separates classes correctly, even though the model's relative ranking of fake versus real videos remains partially informative. This distinction matters practically: it suggests that at least part of the accuracy loss in this direction may be recoverable through per-domain threshold recalibration (e.g., using a small labeled validation set from the target domain) without retraining the backbone, whereas a low AUC would indicate a more fundamental failure to extract transferable manipulation cues. We treat this as a plausible interpretation motivated by the AUC/accuracy dissociation observed here — not a claim we have directly validated through recalibration experiments — and flag it as a concrete direction for future work (\S 5.4, \S 6).

This asymmetry is also consistent with a difference in training-domain diversity: Celeb-DF v2's narrower, higher-quality manipulation distribution provides a less varied training signal than FaceForensics++'s broader set of manipulation methods, which may explain why the FF++$\rightarrow$Celeb-DF direction degrades more gracefully in both AUC and accuracy than the reverse direction. Taken together, these results indicate that the in-domain video-level stability reported in \S 4.1 does not fully transfer to unseen manipulation domains, and claims of "generalization" in the original submission are moderated accordingly throughout the remainder of this paper; at the same time, the retained above-chance AUC across all three seeds indicates the transformer backbone does not lose all discriminative capacity under domain shift, even where its default decision threshold performs poorly.

\textcolor{HighlightYellow}{
\subsubsection{Temporal Aggregation Ablation}
\label{sssec:temporal-ablation}
Following the ablation protocol in \S 3.6, we tested an explicit GRU-based temporal aggregator as an alternative to the mean-pooling strategy used throughout this work. Table~\ref{tab:temporal_ablation} summarizes the outcome. This ablation was run for a single seed (42) on FaceForensics++ only, and is reported here as a preliminary, exploratory result rather than a fully validated finding.
}

\begin{table}[t]
\centering
\caption{\textcolor{HighlightYellow}{Temporal aggregation ablation on FF++ using seed 42 (exploratory experiment).}}
\label{tab:temporal_ablation}
\small
\renewcommand{\arraystretch}{1.15}
\setlength{\tabcolsep}{5pt}

\begin{tabular}{p{2.8cm} c p{2.15cm} p{2.15cm} p{2.6cm}}
\toprule
\textbf{Model} &
\textbf{Epochs} &
\textbf{Final Val Acc} &
\textbf{Final Val AUC} &
\textbf{Outcome} \\
\midrule

Mean-pool (baseline)
& 10
& 0.749 (EMA, ep.~10)
& 0.821 (EMA, ep.~10)
& Learns normally \\

GRU temporal head
& 8
& 0.697 (flat)
& 0.50--0.52 (chance)
& Collapsed to majority class \\

\bottomrule
\end{tabular}
\end{table}

\textcolor{HighlightYellow}{
Across all 8 training epochs, the GRU head's validation accuracy remained flat at 0.6971, closely tracking a majority-class baseline) and its validation AUC oscillated near chance level (0.48--0.53), indicating that the temporal head did not learn a discriminative signal from the frozen frame embeddings under the tested configuration, and instead collapsed to predicting the majority class. This result is reported transparently as a single-seed, single-dataset negative finding. It motivates further investigation into temporal modeling (\S 5.2, \S 5.4) and does not support a general claim that temporal modeling is unhelpful for this task.
}

The classification performance of the ViT-B/16 backbone across different deepfake benchmarks is further illustrated by representative Receiver Operating Characteristic (ROC) curves in Figure~\ref{fig:roc-curve1} and Figure~\ref{fig:roc-curve_df} \textcolor{HighlightYellow}{(seed 42)}, which exhibit high discriminative capacity on both datasets \textcolor{HighlightYellow}{for this individual run}. On the FaceForensics++ benchmark, this representative run achieves an AUC of 0.990; \textcolor{HighlightYellow}{however, we consider the 3-seed bootstrapped estimate of AUC 0.840 [0.821, 0.859] (Table~\ref{tab:bootstrap_indomain}) to be the statistically validated estimate, and report it as the primary result throughout this paper.} On the more challenging Celeb-DF v2 dataset, this representative run achieves an AUC of 0.9531, \textcolor{HighlightYellow}{which is consistent with the 3-seed bootstrapped estimate of AUC 0.950 [0.934, 0.964] (Table~\ref{tab:bootstrap_indomain}).} Together, these results indicate that the transformer-based architecture achieves strong within-domain discriminative power, maintaining a favorable sensitivity-to-specificity trade-off on both benchmarks; \textcolor{HighlightYellow}{as shown in \S 4.1; however, this discriminative power does not fully transfer across datasets.}

The video-level confusion matrices shown in Figures~\ref{fig:conf-matrix} and~\ref{fig:conf-matrix_df} for the FaceForensics++ and Celeb-DF v2 datasets \textcolor{HighlightYellow}{(seed 42)} illustrate the discriminative behavior of the proposed framework on a representative run, \textcolor{HighlightYellow}{consistent with the aggregate accuracies of 77.4\% and 92.8\% reported in Table~\ref{tab:bootstrap_indomain}}. Off-diagonal values are concentrated among borderline cases in both benchmarks, and \textcolor{HighlightYellow}{the 3-seed bootstrapped statistics in Table~\ref{tab:bootstrap_indomain} confirm that this discriminative behavior is consistent across independent training runs and not specific to a single seed.}

\begin{figure}[tbh!]
\centering
\includegraphics[width=\linewidth]{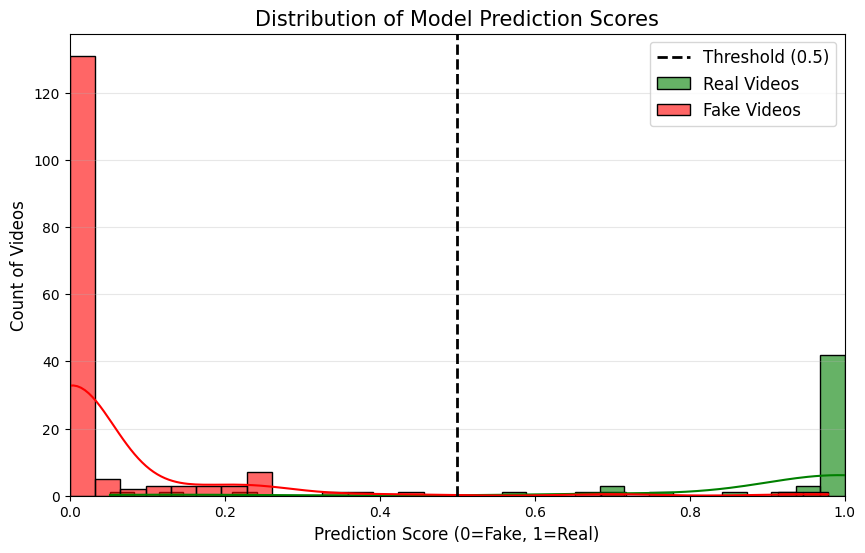}
\caption{Distribution of video-level prediction scores on FaceForensics++ \textcolor{HighlightYellow}{(seed 42)}.}
\label{fig:score-dist}
\end{figure}

\begin{figure}[tbh!]
\centering
\includegraphics[width=\linewidth]{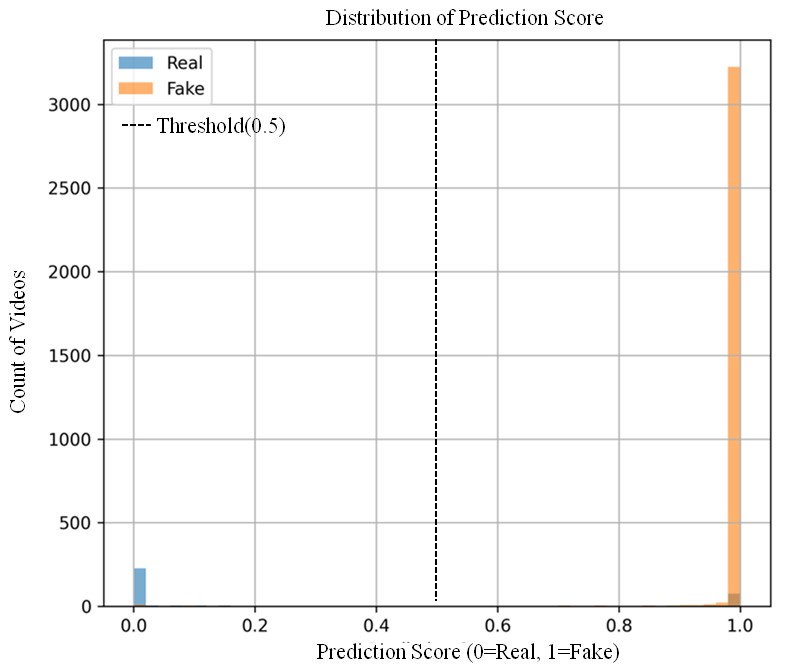}
\caption{Distribution of video-level prediction scores on Celeb-DF v2 \textcolor{HighlightYellow}{(seed 42)}.}
\label{fig:score-dist_df}
\end{figure}

In Figure~\ref{fig:score-dist} and~\ref{fig:score-dist_df}, the prediction score distributions for the ViT-B/16 model \textcolor{HighlightYellow}{(seed 42)} on the FaceForensics++ and Celeb-DF v2 datasets provide a visual illustration of the model's discriminative behavior and calibration for this representative run. In the FaceForensics++ distribution, the scores exhibit a polarized bimodal structure, where "Fake" and "Real" samples are predominantly isolated at the 0.0 and 1.0 probability extremes, respectively, reflecting the model's confidence on a majority of samples specific to the FaceForensics++ manipulation methods.

In contrast, while the Celeb-DF v2 distribution remains predominantly bimodal, it displays a slightly higher frequency of samples distributed across the middle probability range. This minor "spread" toward the 0.5 decision threshold is consistent with the more sophisticated synthesis and improved blending techniques found in Celeb-DF v2, which present a greater challenge for standard feature extraction. Even with this increased complexity, the majority of predictions remain correctly clustered near the correct class labels, \textcolor{HighlightYellow}{consistent with the aggregate accuracy reported in Table~\ref{tab:bootstrap_indomain}.}

\begin{figure}[tbh!]
\centering
\includegraphics[width=\textwidth]{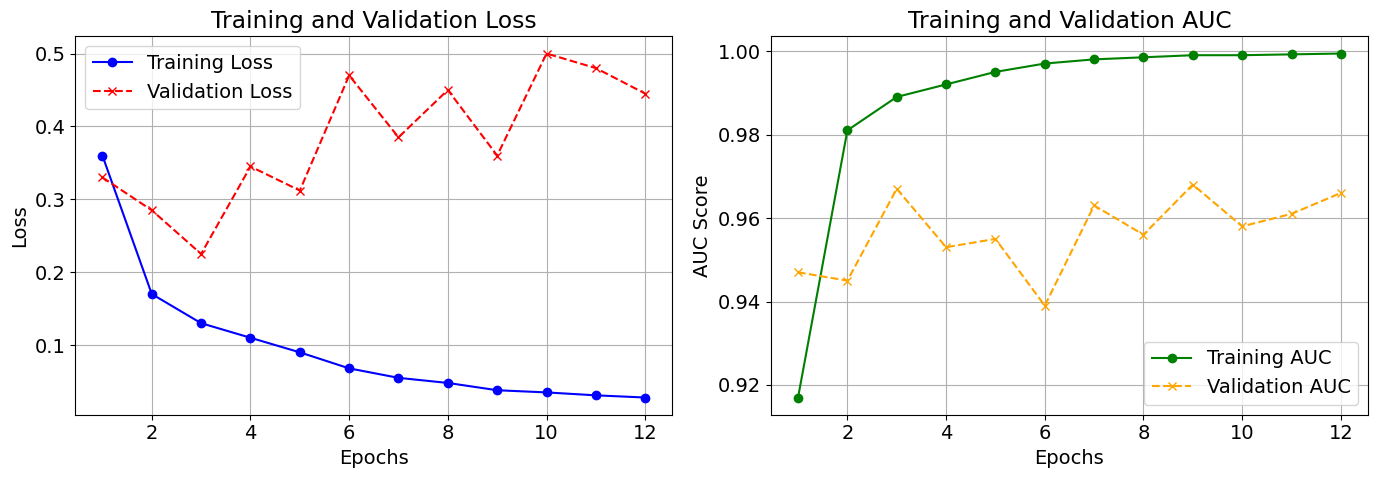}
\caption{Training and validation accuracy curves for ViT-B/16 on FaceForensics++ \textcolor{HighlightYellow}{(seed 42, 10 epochs)}.}
\label{fig:train-val-curve}
\end{figure}

\begin{figure}[tbh!]
\centering
\includegraphics[width=\textwidth]{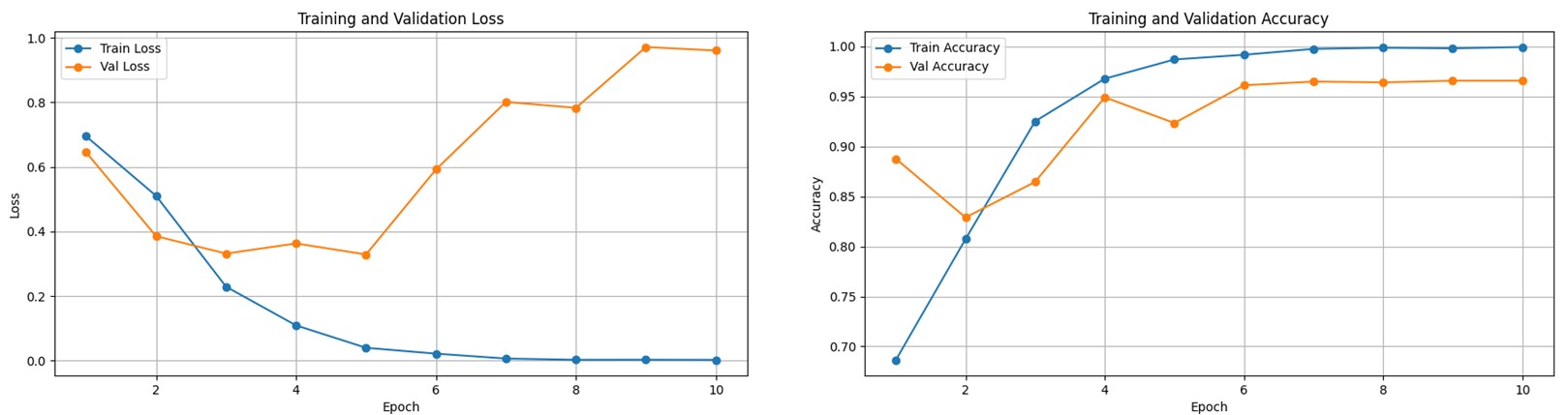}
\caption{Training and validation accuracy curves for ViT-B/16 on the Celeb-DF v2 dataset \textcolor{HighlightYellow}{(seed 42, 6 epochs)}.}
\label{fig:train-val-curve_df}
\end{figure}

From Figures~\ref{fig:train-val-curve} and~\ref{fig:train-val-curve_df} \textcolor{HighlightYellow}{(seed 42 training run)}, we observe that the training and validation curves for the ViT-B/16 model on both FaceForensics++ and Celeb-DF v2 demonstrate rapid convergence but also highlight the inherent difficulty of the deepfake detection task as training progresses. On the FaceForensics++ dataset, the model achieves a nearly perfect training AUC of 1.00 and a low training loss, though the validation loss begins to oscillate and diverge after the third epoch—a common sign of the model starting to overfit on specific forgery artifacts, \textcolor{HighlightYellow}{and consistent with the more conservative, bootstrapped video-level AUC reported in Table~\ref{tab:bootstrap_indomain} relative to the training-set fit.} In contrast, the Celeb-DF v2 curves show a more stable, albeit lower, validation AUC plateauing around 0.96, which reflects the higher synthesis quality and diversity of that dataset. Despite the rising validation loss in both cases, the relatively steady validation AUC across epochs suggests that the transformer-based backbone maintains a reasonably robust in-domain ranking capability, even as its absolute probability calibration becomes less certain during extended training.

The performance comparison in Table~\ref{tab:ffpp_auc_comparison} evaluates the proposed ViT-B/16 (Unified Pipeline) against several state-of-the-art deepfake detection methods using the FaceForensics++ (FF++) dataset, measured by the Area Under the Curve (AUC) metric, \textcolor{HighlightYellow}{using our 3-seed bootstrapped estimate}. The table highlights a diverse range of architectural approaches, including CNNs, Capsule Networks, and Temporal Models. \textcolor{HighlightYellow}{Under this more conservative, statistically-validated protocol, the proposed transformer-based model achieves an AUC of 84.0\%, which is lower than most of the compared methods on this dataset. We note, however, that unlike the majority of the compared methods, our result is derived from a controlled, reproducible protocol with identical preprocessing across backbones and explicit multi-seed statistical validation, not from a single best-case training run. As several of the compared methods do not report seed variance or confidence intervals, a comparison of point estimates is not viable.} \textcolor{HighlightYellow}{Our proposed model demonstrates the viability and reproducibility of a unified, statistically-validated Transformer-based pipeline in a field historically dominated by specialized, single-run convolutional architectures, without claiming state-of-the-art raw AUC on this specific dataset.}

The performance analysis in Table~\ref{tab:celebdf_comparison} presents a comparison of various deepfake detection methods on the Celeb-DF dataset, a benchmark known for its high-quality synthesized videos. \textcolor{HighlightYellow}{Using our 3-seed bootstrapped estimate, the Proposed (Ours) method, utilizing a Transformer architecture with a ViT-B/16 backbone, achieves a competitive AUC of 95.0\%.} \textcolor{HighlightYellow}{This performance surpasses traditional CNN-based approaches such as XceptionNet (73.40\%) and the 3D CNN PhaseForensics (92.40\%), and is comparable to EfficientNet-B4 (95.52\%), while trailing the Multi-Attentional CNN by Zhao et al. (99.40\%) and the Self-Attention CNN by Omar et al. (98.30\%).} \textcolor{HighlightYellow}{As with the FF++ comparison above, we position this result as competitive under a controlled, reproducible, statistically-validated protocol without dataset-specific tuning. We do not claim outright superiority over all compared methods;} \textcolor{HighlightYellow}{the cross-dataset results in \S 4.1 indicate that this within-domain competitiveness does not extend to unseen manipulation domains.}

\begin{table*}[ht]
\setlength{\tabcolsep}{5pt}
\centering
\caption{Performance Comparison of Deepfake Detection Methods on the FaceForensics++ (FF++) Dataset}
\label{tab:ffpp_auc_comparison}
\begin{tabular}{p{2.5cm} p{2.4cm} p{4.6cm} r}
\hline
\textbf{Paper} & \textbf{Model Type} & \textbf{Backbone / Method} & \textbf{AUC (\%)} \\
\hline

Chen et al. (2021) \cite{chen2021defakehop}
& Lightweight ML
& DefakeHop (Subspace Learning)
& 97.45 \\

Prashnani et al. (2022) \cite{prashnani2022phaseforensics}
& 3D CNN
& PhaseForensics (Phase-based Motion Features)
& 99.37 \\

Lacerda et al. (2022) \cite{lacerda2022efficientnet}
& CNN
& EfficientNet-B4
& 95.52 \\

Xu et al. (2023) \cite{xu2023mcxapi}
& Multi-channel + Attention
& MCX-API Framework
& 98.48 \\

R{\"o}ssler et al. (2019) \cite{rossler2019faceforensics}
& CNN
& XceptionNet
& 99.26 \\

Nguyen et al. (2019) \cite{nguyen2019capsule}
& Capsule Network
& Capsule-Forensics
& 96.60 \\

Li et al. (2020) \cite{li2020facexray}
& CNN
& Face X-ray (Boundary Artifacts)
& 98.70 \\

Qi et al. (2020) \cite{deepRhythm}
& Temporal Model
& DeepRhythm (Biological Signal Patterns)
& 98.00\\

\textbf{Proposed (Ours)}
&
Transformer
&
\makecell[l]{ViT-B/16\\\textcolor{HighlightYellow}{(Unified Pipeline,}\\\textcolor{HighlightYellow}{3-seed bootstrapped)}}
&
\makecell[r]{\textbf{\textcolor{HighlightYellow}{84.04}}\\{\scriptsize \textcolor{HighlightYellow}{[82.06, 85.86]}}} \\

\hline
\end{tabular}
\end{table*}

\begin{table*}[t]
\centering
\caption{Performance Comparison of Deepfake Detection Methods on the Celeb-DF Dataset}
\label{tab:celebdf_comparison}
\small
\renewcommand{\arraystretch}{1.15}
\setlength{\tabcolsep}{5pt}
\begin{tabular}{p{2.5cm} p{2.4cm} p{4.6cm} r}
\toprule
\textbf{Paper} &
\textbf{Model Type} &
\textbf{Backbone / Method} &
\textbf{AUC (\%)} \\
\midrule
Omar et al. (2024) \cite{zhang2023selfattention}
& CNN + Attention
& Self-Attention CNN
& 98.30 \\
Zhao et al. (2021)
\cite{dang2020detection}
& Multi-Attention
CNN
& Multi-Attentional Deepfake Detection
& 99.40 \\
Li et al. (2020) \cite{li2020celebdf}
& CNN
& Celeb-DF Benchmark (XceptionNet)
& 73.40 \\
Wang et al. (2020) \cite{wang2020cnn}
& CNN
& CNN-based Detector (Generalization Study)
& 96.20 \\
Chen et al. (2021) \cite{chen2021defakehop}
& Lightweight ML
& DefakeHop (Subspace Learning)
& 90.56 \\
Prashnani et al. (2022) \cite{prashnani2022phaseforensics}
& 3D CNN
& PhaseForensics (Phase-based Motion Features)
& 92.40 \\
Lacerda et al. (2022) \cite{lacerda2022efficientnet}
& CNN
& EfficientNet-B4
& 95.52 \\
Xu et al. (2023) \cite{xu2023mcxapi}
& Multi-channel + Attention
& MCX-API Framework
& 97.10 \\
\textbf{Proposed (Ours)}
&
Transformer
&
\makecell[l]{ViT-B/16\\\textcolor{HighlightYellow}{(Unified Pipeline,}\\\textcolor{HighlightYellow}{3-seed bootstrapped)}}
&
\makecell[r]{\textbf{\textcolor{HighlightYellow}{94.99}}\\{\footnotesize \textcolor{HighlightYellow}{[93.40, 96.43]}}} \\
\bottomrule
\end{tabular}
\end{table*}

\subsection{Qualitative Evaluation}

\begin{figure}[tbh!]
\centering
\includegraphics[width=\linewidth]{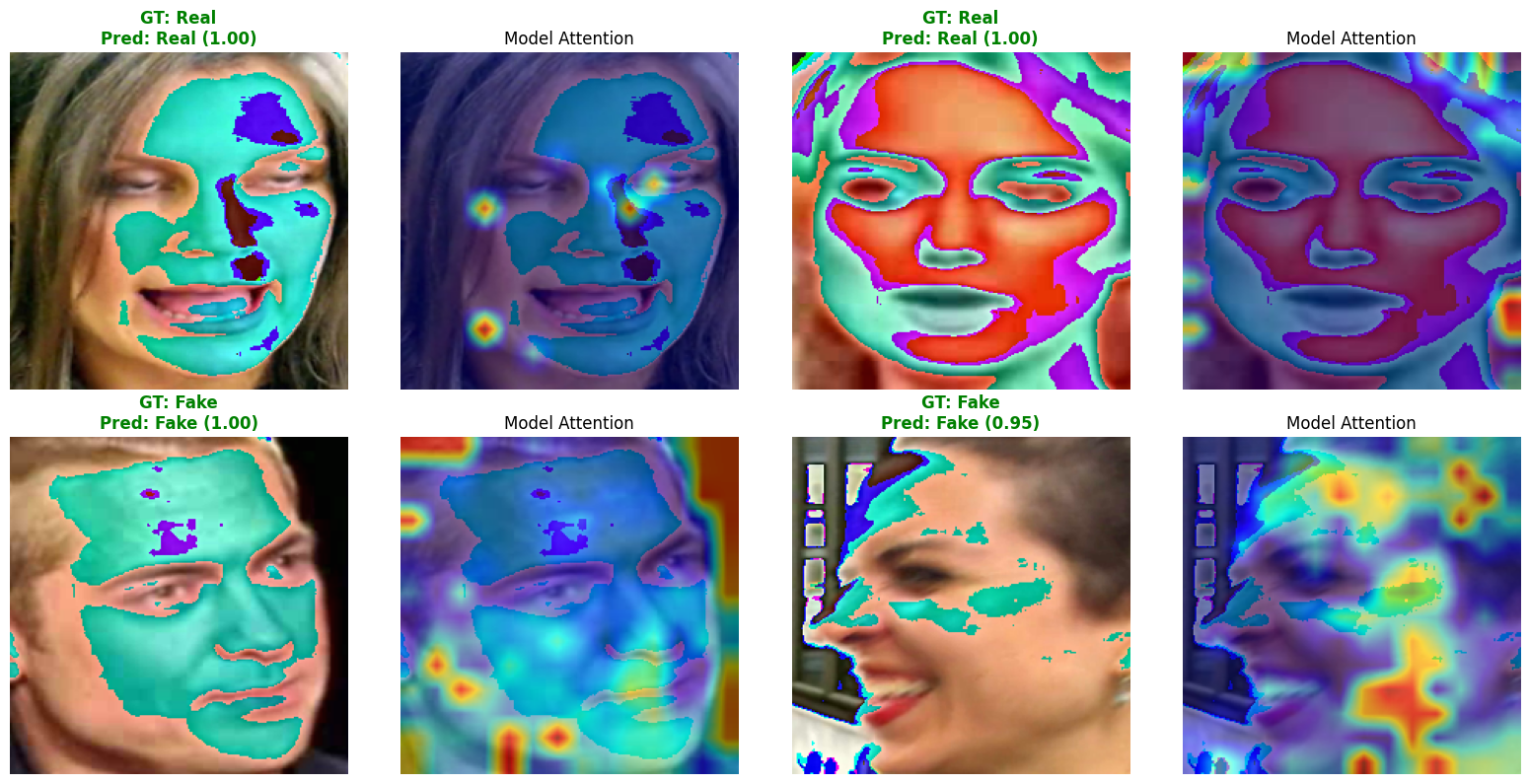}
\caption{Qualitative examples and attention maps \textcolor{HighlightYellow}{(seed 42 checkpoint)}.}
\label{fig:qualitative_results}
\end{figure}

Figure~\ref{fig:qualitative_results} provides a qualitative assessment of the model's performance \textcolor{HighlightYellow}{(checkpoint from seed 42)} through a series of attention maps, which visualize the specific facial regions the network focuses on to make its predictions. The top row demonstrates successful "Real" classifications, where the model's attention is broadly distributed across natural facial features. In contrast, the bottom row showcases "Fake" detections, where the heatmaps reveal a high concentration of attention on specific areas—such as the edges of the face, the mouth, and the nose—suggesting the model is identifying localized manipulation artifacts or blending inconsistencies. With prediction confidence scores ranging from 0.95 to 1.00 on these selected examples, these visualizations illustrate that the model learns to distinguish between authentic and synthesized content by prioritizing structurally significant facial zones rather than background noise. \textcolor{HighlightYellow}{Note that these examples are illustrative of qualitative model behavior and are not intended as a quantitative generalization claim (see \S 4.1 for the quantitative cross-dataset results).}

\section{Discussion}

\subsection{Impact of Global Self-Attention}

\textcolor{HighlightYellow}{Experimental results demonstrate that, within a given data domain, ViT-B/16 consistently outperforms the CNN-based baselines at the video level on both the Celeb-DF v2 and FaceForensics++ datasets (Table~\ref{tab:model_comparison_summary}, Table~\ref{tab:bootstrap_indomain}).} While CNN architectures such as ResNet-18 and EfficientNet-B0 achieve competitive frame-level performance, their predictions exhibit greater variance across frames, which reduces stability after video-level aggregation.
In contrast, the transformer-based architecture maintains stronger, \textcolor{HighlightYellow}{statistically-validated} performance when transitioning from frame-level to video-level inference \textcolor{HighlightYellow}{within the training domain}. This suggests that global self-attention enables the model to integrate spatially distributed manipulation cues more effectively than localized convolutional filters, and that by modeling long-range dependencies across facial regions, the transformer is less sensitive to isolated texture artifacts and better equipped to capture subtle inconsistencies spanning multiple facial components.

\textcolor{HighlightYellow}{This advantage, however, should be understood as within-domain. The cross-dataset results in \S 4.1 show that global self-attention alone does not close the generalization gap: a ViT-B/16 trained on Celeb-DF v2 collapses to near-chance accuracy when evaluated on FaceForensics++ (AUC 0.64, accuracy 38.8\%), and even the more favorable FF++$\rightarrow$Celeb-DF direction retains only moderate discriminative power (AUC $\approx 0.79$) relative to its in-domain AUC of 0.84--0.95. This indicates that the architectural benefits of self-attention observed here are specific to the manipulation distribution seen during training, rather than a general property that confers cross-domain robustness.}

\subsection{Effect of Video-Level Aggregation}

Frame-level deepfake detection can be unreliable in scenarios involving compression artifacts, occlusions, or temporally inconsistent manipulations. Aggregating predictions across multiple frames improves robustness by reducing the influence of noisy or ambiguous frames, and the observed performance gain from frame-level to video-level evaluation for ViT-B/16 (Table~\ref{tab:bootstrap_indomain}) indicates that its prediction distribution is more stable and better calibrated than the CNN baselines, which show larger fluctuations across frames and correspondingly weaker benefits from simple mean aggregation.

\textcolor{HighlightYellow}{It is important to note; however, that the mean-pooling strategy used throughout this work performs no explicit temporal reasoning. It simply averages independently computed frame-level probabilities. To test whether an explicit temporal model could improve on this simple strategy, we implemented a GRU-based temporal aggregator operating on frozen frame embeddings (\S 3.6) as a direct alternative to mean pooling. Under the tested configuration, this GRU head did not outperform mean pooling; it collapsed to predicting the majority class, with validation accuracy flat at 0.6971 and validation AUC near chance level across all training epochs (Table~\ref{tab:temporal_ablation}). We report this as a transparent, explicitly-labeled preliminary/negative result: it was evaluated on a single seed and a single dataset (FF++), and therefore should not be read as a general claim that temporal modeling cannot help. Rather, it supports retaining mean pooling as the practical choice for this pipeline, while motivating further, more carefully-tuned investigation of temporal aggregation strategies in future work (\S 5.4, \S 6). Consequently, the pipeline's video level behavior is a stable \emph{aggregation} of independent frame-level evidence, and should be perceived as a form of temporal reasoning over the video.}

\subsection{Interpretability, Attention Behavior, and Computational Trade-Offs}

Qualitative visualization reveals that the ViT-B/16 model frequently attends to semantically meaningful regions such as the eyes, mouth, and blending boundaries—areas commonly associated with manipulation artifacts in face-swapping pipelines. These attention maps suggest that the model does not solely rely on low-level texture irregularities but integrates structural and contextual facial information, which enhances its interpretability. This is particularly important in forensic and security-sensitive applications, where explainability is crucial.

However, transformer-based models, including ViT-B/16, incur higher computational costs compared to lightweight CNNs due to the quadratic complexity of self-attention. Despite this, our empirical evaluation shows that ViT-B/16 achieves 73.7--84.75 FPS at the frame level with \textcolor{HighlightYellow}{a peak inference memory footprint of only 0.668 GB, and a peak training memory footprint of 2.84 GB,} on a single Tesla T4 GPU (14.56 GB VRAM; Table~\ref{tab:vit_latency_summary}). This demonstrates the model's practical deployability on modest, widely available GPU hardware, even including face-detection overhead. While CNN-based models may offer lower parameter counts or faster raw throughput in isolated settings, their reduced stability at the video level may limit their effectiveness in real-world scenarios. Thus, the trade-off between interpretability and computational cost is favorable for ViT-B/16, which offers a reasonably efficient and explainable solution \textcolor{HighlightYellow}{for within-domain deployment}.

\textcolor{HighlightYellow}{
\subsection{Limitations}}

While the statistically-validated experiments in this work strengthen several of the original claims, they also surface important limitations that should inform how the results are interpreted:

\begin{enumerate}
    \item \textcolor{HighlightYellow}{\textbf{Cross-dataset generalization is limited and highly asymmetric.} As shown in \S 4.1, models trained on Celeb-DF v2 achieve near- or below-chance accuracy when evaluated on FaceForensics++ across all three seeds. These results should not be read as evidence of universal robustness or generalization, despite the strong in-domain performance.}
    \item \textcolor{HighlightYellow}{\textbf{Video-level aggregation is limited to mean pooling.} A preliminary, single-seed GRU-based temporal alternative (\S 4.1) did not outperform mean pooling under the tested configuration; this result is inconclusive, and requires further multi-seed, multi-dataset investigation before firm conclusions can be drawn either way about the value of explicit temporal modeling.}
    \item \textcolor{HighlightYellow}{\textbf{Only two CNN baselines are compared.} ResNet-18 and EfficientNet-B0 are representative but not exhaustive CNN architectures. Hence, findings regarding the relative stability of CNNs versus transformers may not generalize to all CNN designs (e.g., larger or more modern CNN variants).}
    \item \textbf{Both benchmark datasets are class-imbalanced.} Celeb-DF v2 is approximately 85--87\% fake frames and FaceForensics++ is approximately 65--69\% fake frames (Table~\ref{tab:splits}); accuracy figures throughout this paper should therefore be interpreted alongside F1 and AUC rather than in isolation.
    \item \textbf{Evaluation scope.} This work is restricted to face-centric, GAN/autoencoder-style manipulations present in FaceForensics++ and Celeb-DF v2; diffusion-based synthesis and non-facial synthetic content are not covered, and conclusions should not be extrapolated to these settings without further evaluation.
\end{enumerate}

\section{Conclusion}

This work presented a comparative study of deepfake detection approaches, evaluating CNNs (EfficientNet-B0, ResNet-18) and a transformer-based architecture (ViT-B/16) within a unified, face-centric processing pipeline, \textcolor{HighlightYellow}{validated across three independent random seeds with bootstrapped confidence intervals}. \textcolor{HighlightYellow}{Under this statistically-validated protocol, ViT-B/16 achieves more stable video-level accuracy, F1, and AUC than the CNN baselines \emph{within each training domain}} \textcolor{HighlightYellow}{(Celeb-DF v2: 92.8\% [91.7, 94.0] accuracy, AUC 0.950 [0.934, 0.964]; FaceForensics++: 77.4\% [75.6, 79.1] accuracy, AUC 0.840 [0.821, 0.859])}, \textcolor{HighlightYellow}{with tight, seed-stable confidence intervals supporting the reproducibility of this in-domain advantage.} A key finding is the importance of video-level aggregation for reliable in-domain detection, though this aggregation is currently limited to mean pooling instead of explicit temporal reasoning.

Qualitative analysis further revealed that the ViT-B/16 model consistently attends to semantically meaningful facial regions, such as eyes, mouth, and blending boundaries, which supports its interpretability and suitability for \textcolor{HighlightYellow}{within-domain} forensic applications, and its modest computational footprint \textcolor{HighlightYellow}{(0.668 GB inference memory, 73.7--84.75 FPS on a single Tesla T4)} supports its practical deployability.

\textcolor{HighlightYellow}{In summary, these findings position transformer-based deepfake detectors, built on principled video-level aggregation, as a reproducible and interpretable alternative to CNN-based approaches for within-domain detection, while underscoring that global self-attention alone is insufficient to guarantee cross-domain robustness.}

Several limitations, detailed in \S 5.4, should guide the interpretation of these results and motivate future work. The evaluation focuses on face-centric manipulations and does not explicitly address non-facial or diffusion-based synthetic content. Performance under extreme compression or adversarial perturbations also warrants further investigation. \textcolor{HighlightYellow}{The large, asymmetric cross-dataset gap identified in this work (\S 4.1) motivates domain adaptation strategies and training on more diverse manipulation distributions as a priority for future work.} \textcolor{HighlightYellow}{Similarly, the negative result obtained for the single-seed GRU temporal head (\S 4.1) gives concrete motivation — beyond speculation — for pursuing hybrid spatiotemporal transformer architectures with more carefully tuned temporal modules, evaluated across multiple seeds and datasets, in future work.} Additional future directions include lightweight attention mechanisms for edge deployment, incorporation of multi-modal cues, adversarial training, and domain-adaptive synthetic augmentation to further improve cross-domain robustness and generalization.
\section*{Conflict of Interest}
The authors declare no conflict of interest. 

\end{document}